\documentclass[preprint,12pt]{elsarticle}

\usepackage{amssymb}
\usepackage{multirow}
\usepackage{color}
\usepackage{caption}
\usepackage{subcaption}
\usepackage{colortbl}
\usepackage{xspace}
\usepackage{hyperref}
\newcommand{\ie}{{\emph{i.e.}},\xspace}

\journal{}

\begin{document}

\begin{frontmatter}



\title{SGDFormer: One-stage Transformer-based Architecture for Cross-Spectral Stereo Image Guided Denoising}

\author[label1]{Runmin Zhang}
\ead{runmin_zhang@zju.edu.cn}

\author[label1]{Zhu Yu}
\ead{yu_zhu@zju.edu.cn}

\author[label1]{Zehua Sheng}
\ead{shengzehua@zju.edu.cn}

\author[label1]{Jiacheng Ying\corref{correspondingauthor}}
\cortext[correspondingauthor]{Corresponding author}
\ead{yingjiacheng@zju.edu.cn}

\author[label2,label1]{Si-Yuan Cao}
\ead{cao_siyuan@zju.edu.cn}

\author[label3]{Shu-Jie Chen}
\ead{chenshujie@zjgsu.edu.cn}

\author[label3]{Bailin Yang}
\ead{ybl@zjgsu.edu.cn}

\author[label1,label4]{Junwei Li}
\ead{lijunwei7788@zju.edu.cn}

\author[label1,label5]{Hui-Liang Shen\corref{correspondingauthor}}
\ead{shenhl@zju.edu.cn}

\affiliation[label1]{organization={College of Information Science and Electronic Engineering, Zhejiang University}, 
	city={Hangzhou},
	postcode={310027}, 
	state={Zhejiang},
	country={China}}

\affiliation[label2]{organization={Ningbo Research Institute, Zhejiang University},
	city={Ningbo},
	postcode={315100},
	state={Zhejiang},
	country={China}}

\affiliation[label3]{organization={Zhejiang GongShang University},
	addressline={JiaoGong Road 149}, 
	city={Hangzhou},
	postcode={310012}, 
	country={China}}

\affiliation[label4]{organization={Key Laboratory of Collaborative Sensing and Autonomous Unmanned Systems of Zhejiang Province}, 
	city={Hangzhou},
	postcode={310015}, 
	country={China}}

\affiliation[label5]{organization={Jinhua Institute, Zhejiang University}, 
	city={Jinhua},
	postcode={321299}, 
	country={China}}

\begin{abstract}
Cross-spectral image guided denoising has shown its great potential in recovering clean images with rich details, such as using the near-infrared image to guide the denoising process of the visible one. To obtain such image pairs, a feasible and economical way is to employ a stereo system, which is widely used on mobile devices. Current works attempt to generate an aligned guidance image to handle the disparity between two images. However, due to occlusion, spectral differences and noise degradation, the aligned guidance image generally exists ghosting and artifacts, leading to an unsatisfactory denoised result. To address this issue, we propose a one-stage transformer-based architecture, named SGDFormer, for cross-spectral \textbf{S}tereo image \textbf{G}uided \textbf{D}enoising. The architecture integrates the correspondence modeling and feature fusion of stereo images into a unified network. Our transformer block contains a noise-robust cross-attention (NRCA) module and a spatially variant feature fusion (SVFF) module. The NRCA module captures the long-range correspondence of two images in a coarse-to-fine manner to alleviate the interference of noise. The SVFF module further enhances salient structures and suppresses harmful artifacts through dynamically selecting useful information. Thanks to the above design, our SGDFormer can restore artifact-free images with fine structures, and achieves state-of-the-art performance on various datasets. Additionally, our SGDFormer can be extended to handle other unaligned cross-model guided restoration tasks such as guided depth super-resolution.
\end{abstract}

%


\end{frontmatter}


\section{Introduction}
\label{sec:intro}

Image denoising is a fundamental task in the field of image processing. However, due to the inherent ill-condition issue, the restoration result of single image denoising methods~\cite{ReviewDenoising_IF20, DnCNN_TIP17, MPRNet_CVPR21, NAFNet_ECCV22, MalleNet_ECCV22} may be over-smoothed and lose details, especially in the case of high noise levels.

To address this problem, guided denoising~\cite{GF_TPAMI12, DarkVisionNet_AAAI22, FGDNet_TMM22, MNNet_IF22, SANet_CVPR23}, which uses another noise-free guidance image as extra information source, has been introduced. It aims to transfer the structural details of the guidance image into the restored image while preserving the color and brightness of the target noisy image. For instance, by integrating an additional invisible near-infrared (NIR) light source, we can obtain clean NIR images to guide the denoising process of RGB images~\cite{DarkVisionNet_AAAI22, Splitter_CVPR08, Splitter_ECCV20, Rotator_AAAI20, SDF_ICCP19}.

Most of the current guided denoising approaches~\cite{DarkVisionNet_AAAI22, FGDNet_TMM22, MNNet_IF22} assume that the guidance image and the target image are strictly aligned, thus the specific equipment should be embedded into the camera system to ensure this assumption~\cite{Splitter_CVPR08, Splitter_ECCV20, Rotator_AAAI20, Mutually_IF23}. These specialized designs are too complex to be deployed on mobile devices, limiting the practical application of guided denoising algorithms. A more cost-effective method is to use a stereo system for multi-spectral image acquisition~\cite{SDF_ICCP19, SANet_CVPR23}.

To deal with the disparity between stereo image pairs, a straightforward way is to employ the stereo or optical flow algorithms to align input images~\cite{SDF_ICCP19, MRJF_ICCV17, SPIMNet_IF23}. Nevertheless, the optimal motion estimation cannot guarantee the most appropriate guidance image. To alleviate this problem, SANet~\cite{SANet_CVPR23} estimates a structure map as the aligned guidance image by aggregating non-local pixel values within the maximum disparity. However, the aforementioned approaches model the stereo guided denoising process by the two separate steps: aligned guidance image generation and guided denoising. The guidance image generation and the guided denoising are decoupled, and solely trained with the respective supervision. As the ghosting and artifacts are inevitable due to occlusion, spectral differences and noise degradation, the guided denoising step has to tolerate the undesired guidance produced by the former guidance image generation step, leading to unsatisfactory denoised results. 

\begin{figure}[t]
	\centering
	\includegraphics[scale=0.5] {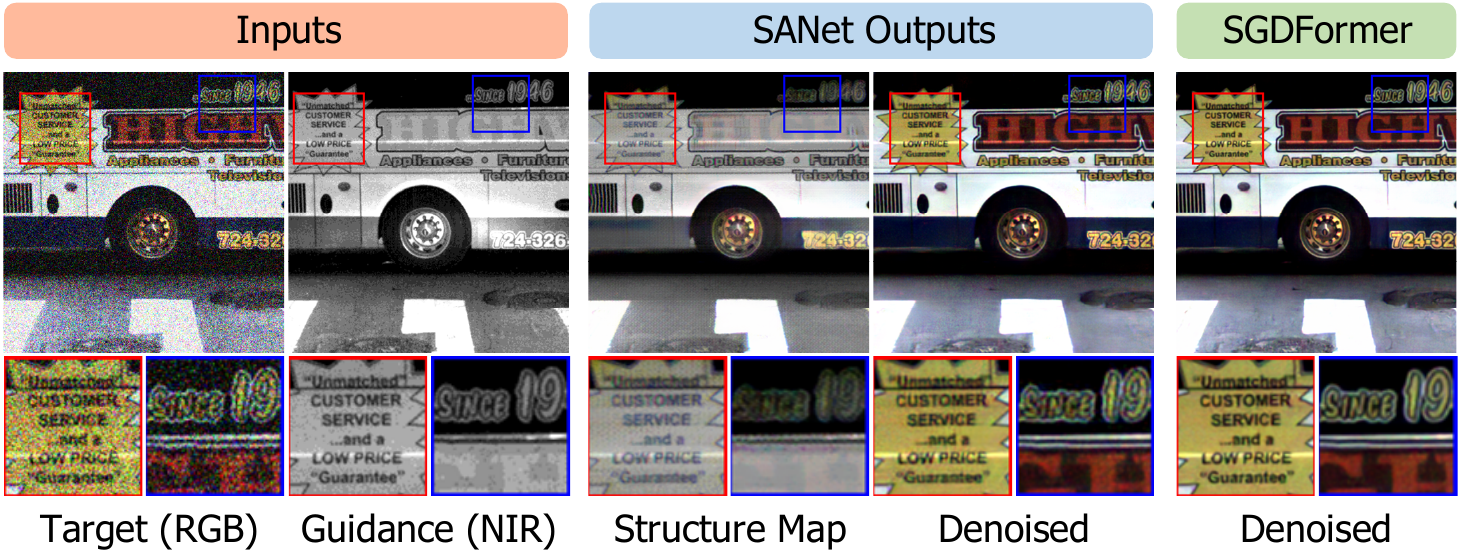}
	\caption{Comparison between the previous state-of-the-art approach SANet~\cite{SANet_CVPR23} and our SGDFormer. SANet separates stereo guided denoising into two steps: aligned guidance image generation and guided denoising. The latter step has to tolerate the undesired guidance (structure map) estimated by the former, generally leading to the unsatisfactory denoised image. In contrast, our SGDFormer integrates the correspondence modeling and feature fusion of two images into a one-stage architecture. In this way, information of the guidance image is preserved to the best extent, thus effectively removing noise while restoring fine structures.}
	\label{fig:Illustration}
\end{figure}

To cope with the above problems, in this work, we propose a specifically designed transformer-based architecture for cross-spectral \textbf{S}tereo image \textbf{G}uided \textbf{D}enoising, named SGDFormer, which directly models the long-range correspondence between two images and then performs feature fusion. By avoiding explicit image registration or aligned guidance image generation, our SGDFormer can be trained in a one-stage end-to-end manner. In this way, the original information of the guidance image is preserved to the best extent. Better still, the correspondence modeling and feature fusion are absorbed into a unified architecture, enabling the mutual promotion of the two parts. Different from common transformers, our SGDFormer are composed of the noise-robust cross-attention (NRCA) module and the spatially variant feature fusion (SVFF) module. As the target image is severely corrupted by noise, directly computing the attention map on the pixel-level may cause inaccurate correspondence. To deal with the problem, the NRCA module captures the long-range correspondence of two images in a coarse-to-fine manner, based on the basic idea that the local consistency of the correspondence can alleviate the interference of noise. Furthermore, the SVFF module performs feature fusion of two images by predicting spatially variable weights according to the content of features. Under a unified transformer architecture, the feature fusion strategy can better enhance rich details and further suppresses harmful artifacts of input features. Extensive comparison experiments are conducted to illustrate the effectiveness of our SVFF module. 

Benefiting from the above designs, our SGDFormer significantly outperforms previous approaches on various datasets. As illustrated in Fig.~\ref{fig:Illustration}, compared with the previous state-of-the-art approach SANet~\cite{SANet_CVPR23}, our SGDFormer can restore artifact-free denoised images with more salient structures. Moreover, our SGDFormer also has the potential to cope with other unaligned cross-model guided restoration tasks such as guided depth super-resolution. Our main contributions can be summarized as follows:

\begin{itemize}
	\item[$\bullet$] We propose a one-stage transformer-based architecture for cross-spectral stereo image guided denoising, named SGDFormer, which integrates the correspondence modeling and feature fusion into a unified network. By eliminating the explicit generation of the aligned guidance, our SGDFormer can better utilize the information of the guidance image, resulting in the artifact-free denoised image.
	\item[$\bullet$] We devise the noise-robust cross-attention (NRCA) module and the spatially variant feature fusion (SVFF) module to constitute the unified transformer architecture. Specifically, the NRCA module captures the long-range correspondence of two images in a coarse-to-fine manner to mitigate the interference of noise. The SVFF module employs a simple but effective spatially variant fusion strategy to further enhance structures and suppress harmful artifacts.
	\item[$\bullet$] Extensive experiments demonstrate our method achieves the state-of-the-art performance on both synthetic and real-world datasets. More experimental results show that our method can be extended to handle other unaligned cross-model guided restoration tasks such as guided depth super-resolution.
\end{itemize}

\section{Related Work}
\label{sec:related}

\subsection{Single Image Denoising}

Single image denoising aims to recover the clean image from its degraded version. Traditional denoising approaches are usually solved in a mathematical optimization problem regularized by the image priors~\cite{ReviewDenoising_IF20}, such as non-local similarity~\cite{NLMeans_CVPR05, BM3D_TIP07}, low-rankness~\cite{Lowrank_TCSVT17}, and sparsity~\cite{Sparsity_TIP06}. As a consequence, these approaches require a long computational time, and cannot handle complex noise distributions.

In recent years, CNN-based networks have been widely employed to improve denoising performance. DnCNN~\cite{DnCNN_TIP17} uses a deep CNN with residual learning and batch normalization for Gaussian denoising. MPRNet~\cite{MPRNet_CVPR21} builds a multi-stage architecture and injects supervision at each stage to progressively restore the degraded image. DGUNet~\cite{DGUNet_CVPR22} unfolds the proximal gradient descent algorithm into a network to combine merits of traditional optimization and deep learning. MalleNet~\cite{MalleNet_ECCV22} adopts the malleable convolution to fit the spatially varying patterns in natural images. NAFNet~\cite{NAFNet_ECCV22} forms a nonlinear activation free network for image restoration, which is computationally efficient. In addition, vision transformers~\cite{VIT} have shown its superior performance in image denoising, due to the long-range modeling ability. IPT~\cite{IPT_CVPR21} first develops a pre-trained model using transformer architecture for image restoration. To reduce the computational complexity, Uformer~\cite{Uformer_CVPR22} presents a U-Net based network with locally-enhanced window transformer blocks, while Restormer~\cite{Restormer_CVPR22} applies self-attention across channels rather than spatial dimensions. To mitigate the artifacts and missing of textual details in the denoised images, EFF-Net~\cite{EFFNet_IF23} introduces an adaptive frequency enhancement transformer block to selectively recovery different frequencies through the long-range dependency. Moreover, CTNet~\cite{CTNet_IF24} embeds transformer mechanisms into a CNN architecture to extract salient features for removing noise. However, it is still difficult for single image denoising approaches to restore rich details due to the ill-posed nature, especially at high noise levels.

\subsection{Guided Image Denoising}
Guided image denoising introduces another guidance image to guide the denoising process of the noisy target image, aiming to preserve the structures and details of the image while denoising. Guided image filter~\cite{GF_TPAMI12} assumes that the filtering output can be linearly represented by the guidance image in local patches. To address the structure inconsistency problem between input images, the work~\cite{ScaleMap_ICCV13} explicitly generates a scale map to preserve the necessary edges and details for visually compelling image restoration. Furthermore, the work~\cite{MSJF_ICCV15} introduces the concept of mutual-structure to distinguish the structural information contained in both images.

Recently, deep learning has been an important tool for guided denoising. DJF~\cite{DJF_ECCV16} introduces a learning-based joint filter utilizing CNNs, and SVLRM~\cite{SVLRM_CVPR19} builds a spatially variant linear representation model with learnable coefficients. Based on the convolutional sparse coding model, CU-Net~\cite{CUNet_TPAMI20} splits image features into common ones and unique ones. Further, FGDNet~\cite{FGDNet_TMM22} introduces frequency decomposition into guided image denoising, while MNNet~\cite{MNNet_IF22} formulates an deep implicit prior regularized optimization problem. However, these methods assume that the target and guidance images are well registered.

To address the misalignment of image pairs, MRGF~\cite{MRJF_ICCV17} generates a set of translated guidances for joint filter. Some works~\cite{SDF_ICCP19, DRF_CVPR21} directly employ existing flow estimation algorithms~\cite{BF_TOG16, PWCNet_CVPR18} to register the image pairs, and then perform aligned guided denoising. Morever, SPIMNet~\cite{SPIMNet_IF23} introduces a domain translation network and two siamese feature extraction networks for cross-spectral image matching. Given the stereo pairs of noisy RGB and clean NIR images, SPIMNet first aligns the image pairs and then denoises the RGB image through joint filter. To avoid the use of ground-truth motion field, SANet~\cite{SANet_CVPR23} uses a structure aggregation module to estimate a structure map from the input stereo pairs as guidance. Since the above approaches separate aligned guidance generation and guided denoising into two steps, the denoised images generally suffer from artifacts.

Different from the above works, our SGDFormer is a one-stage architecture for cross-spectral stereo image guided denoising. Specifically, the transformer blocks of SGDFormer directly utilize the non-local information of the guidance image to guide the denoising process of the noisy target image. Thanks to its unified architecture, our network can be trained end-to-end, thus preserving the information from the guidance image to the best extent.

\subsection{Stereo Image Restoration}
Stereo image restoration aims to take advantage of the redundant information between the cross-view images for structure recovery. DAVANet~\cite{DAVANet_CVPR19} uses the pridicted disparity to aggregate features between the left and right images for stereo deblurring. Besides, StereoIRN~\cite{StereoIRN_CVPR20} builds a unified framework for both stereo image restoration and disparity estimation. However, these approaches require ground-truth disparities for supervision. To capture stereo correspondence without disparity supervision, PASSRnet~\cite{PASSRnet_CVPR19} introduces the parallax-attention mechanism that computed the feature similarities along the epipolar line. Furthermore, SIR-Former~\cite{SIR-Former_ACMMM22} adopts a transformer architecture for feature alignment and fusion. NAFSSR~\cite{NAFSSR_CVPRW22} extends NAFNet~\cite{NAFNet_ECCV22} to stereo image super-resolution through stereo cross-attention. Recently, ACLRNet~\cite{ACLRNet_TPAMI24} introduces an attention-guided correspondence learning network to learn both self-view and cross-view correspondence through parallax and ominidirectional attention.

Different from above stereo restoration approaches which capture correspondence between two RGB images, cross-spectral stereo image guided denoising needs to establish the correspondence between two images under large degradation and spectral differences. To obtain more reliable correspondence, our proposed NRCA module performs in a coarse-to-fine manner and takes the local consistency into consideration, which is different from previous attention mechanism~\cite{PASSRnet_CVPR19, SIR-Former_ACMMM22, NAFSSR_CVPRW22, ACLRNet_TPAMI24}.

\section{Method}
\label{sec:method}

\subsection{Preliminaries}
Let $\mathbf{I}_\mathrm{T}$ and $\mathbf{I}_\mathrm{G}$ denote the target image and the guidance image captured by the stereo system. We assume that $\mathbf{I}_\mathrm{T}$ and $\mathbf{I}_\mathrm{G}$ are captured in the left and the right view respectively, and there are horizontal disparities between the image pairs~\cite{Stereo_Survey0, Unify_TPIMI23}, with maximal disparity $D$. $\mathbf{I}_\mathrm{T}$ is corrupted by the additive noise $\mathbf{N}$, \ie $\mathbf{I}_\mathrm{N} = \mathbf{I}_\mathrm{T} + \mathbf{N}$, where $\mathbf{I}_\mathrm{N}$ denotes the noisy observation of the target image.

The key of stereo guided denoising is to search and aggregate the non-local information from $\mathbf{I}_\mathrm{G}$ that contributes to the noise reduction and structure recovery of $\mathbf{I}_\mathrm{N}$. The process can be formulated as
\begin{equation}
	\mathbf{I}_\mathrm{R} = \mathcal{G}(\mathbf{I}_\mathrm{N}, \mathcal{A}(\mathbf{I}_\mathrm{G})),
	\label{eq:SGD}
\end{equation}
where $\mathcal{A}(\cdot)$ denotes the non-local information aggregation, $\mathcal{G}(\cdot, \cdot)$ denotes the guided denoising process, and $\mathbf{I}_\mathrm{R}$ is the denoised image. For the stereo scene, the searching region of $\mathcal{A}(\cdot)$ is set to the epipolar line within $D$. Previous approaches~\cite{SDF_ICCP19, SANet_CVPR23} treat $\mathcal{A}(\cdot)$ and $\mathcal{G}(\cdot, \cdot)$ as separate issues, resulting in inadequate utilization of $\mathbf{I}_\mathrm{G}$ and artifacts of $\mathbf{I}_\mathrm{R}$. To better utilize the information of $\mathbf{I}_\mathrm{G}$, we propose SGDFormer, a one-stage transformer-based architecture which directly models the correspondence between two images and then performs feature fusion, without explicit aligned guidance image generation.

As the guided denoising process is identical for each channel of the target image, we assume $\mathbf{I}_\mathrm{N}, \mathbf{I}_\mathrm{G} \in \mathbb{R}^{H\times W}$, where $H$ and $W$ denote the height and width of the image. The network processes one channel at a time. For instance, the R, G, and B channels of the color image are denoised separately under the guidance of the NIR channel. 

\subsection{Overall Architecture}

\begin{figure*}[t]
	\centering
	\includegraphics[scale=0.75]{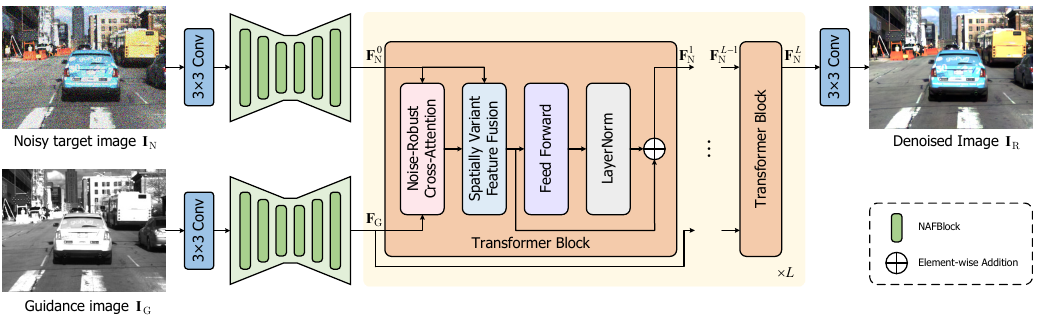}
	\caption{The overall architecture of our stereo guided denoising network SGDFormer.}
	\label{fig:Method}
\end{figure*}

Fig.~\ref{fig:Method} shows the overall architecture of SGDFormer. Given two images $\mathbf{I}_\mathrm{N}$ and $\mathbf{I}_\mathrm{G}$, we first use a $3\times3$ convolution layer to extract shallow features, and then employ an U-shaped~\cite{Unet} feature extractor to extract deep features $\mathbf{F}^{0}_\mathrm{N}$, $\mathbf{F}_\mathrm{G} \in \mathbb{R}^{H\times W\times C}$, where $C$ denotes the feature dimension. The feature extractor has three scales, and uses the NAFBlock~\cite{NAFNet_ECCV22} as the basic block. The parameters of two feature extractors are not shared due to the degradation and spectral differences between two input images.

Then, we stack $L$ transformer blocks to perform feature interaction of two images. The transformer block mainly consists of a noise-robust cross-attention (NRCA) module and a spatially variant feature fusion (SVFF) module. The former models the stereo correspondence between two images, and the latter performs feature fusion of two images. Specifically, given the input features $\mathbf{F}^{l-1}_\mathrm{N}$ and $\mathbf{F}_\mathrm{G}$, the process of the $l$-th transformer block is formulated as:
\begin{equation}
	\mathbf{F}'^{l}_\mathrm{G} = \mathtt{NRCA}(\mathbf{F}^{l-1}_\mathrm{N}, \mathbf{F}_\mathrm{G}),
	\label{eq:transformer_block0}
\end{equation}
\begin{equation}
	\mathbf{F}^{l}_\mathrm{R} = \mathtt{SVFF}(\mathbf{F}^{l-1}_\mathrm{N}, \mathbf{F}'^{l}_\mathrm{G}),
	\label{eq:transformer_block1}
\end{equation}
\begin{equation}
	\mathbf{F}^{l}_\mathrm{N} = \mathbf{F}^{l}_\mathrm{R} + \mathtt{LN}(\mathtt{FFN}(\mathbf{F}^{l}_\mathrm{R})),
	\label{eq:transformer_block2}
\end{equation}
where $\mathbf{F}'^{l}_\mathrm{G}$ is the output of the NRCA module, $\mathbf{F}^{l}_\mathrm{R}$ is the output of the SVFF module, FFN denotes the feed-forward network, and LN denotes the layer normalization.

Finally, we estimate the denoised image $\mathbf{I}_\mathrm{R}$ from the output features $\mathbf{F}^{L}_\mathrm{N}$ of the last transformer block through a $3\times3$ convolution layer. With the above designs, the correspondence modeling and feature fusion are integrated into a unified architecture. As the network is trained end-to-end, the two components can mutually promote each other, resulting in better denoised images.

We train the network by minimizing the PSNR loss~\cite{HINet_CVPRW21, NAFNet_ECCV22} between $\mathbf{I}_\mathrm{R}$ and the ground-truth clean image $\mathbf{I}_\mathrm{T}$. The loss function is defined as
\begin{equation}
	\mathcal{L} = -\mathtt{PSNR}(\mathbf{I}_\mathrm{R}, \mathbf{I}_\mathrm{T}).
	\label{eq:loss}
\end{equation}

\subsection{Noise-Robust Cross-Attention (NRCA)}

\begin{figure}[t]
	\centering\includegraphics[scale=0.24]{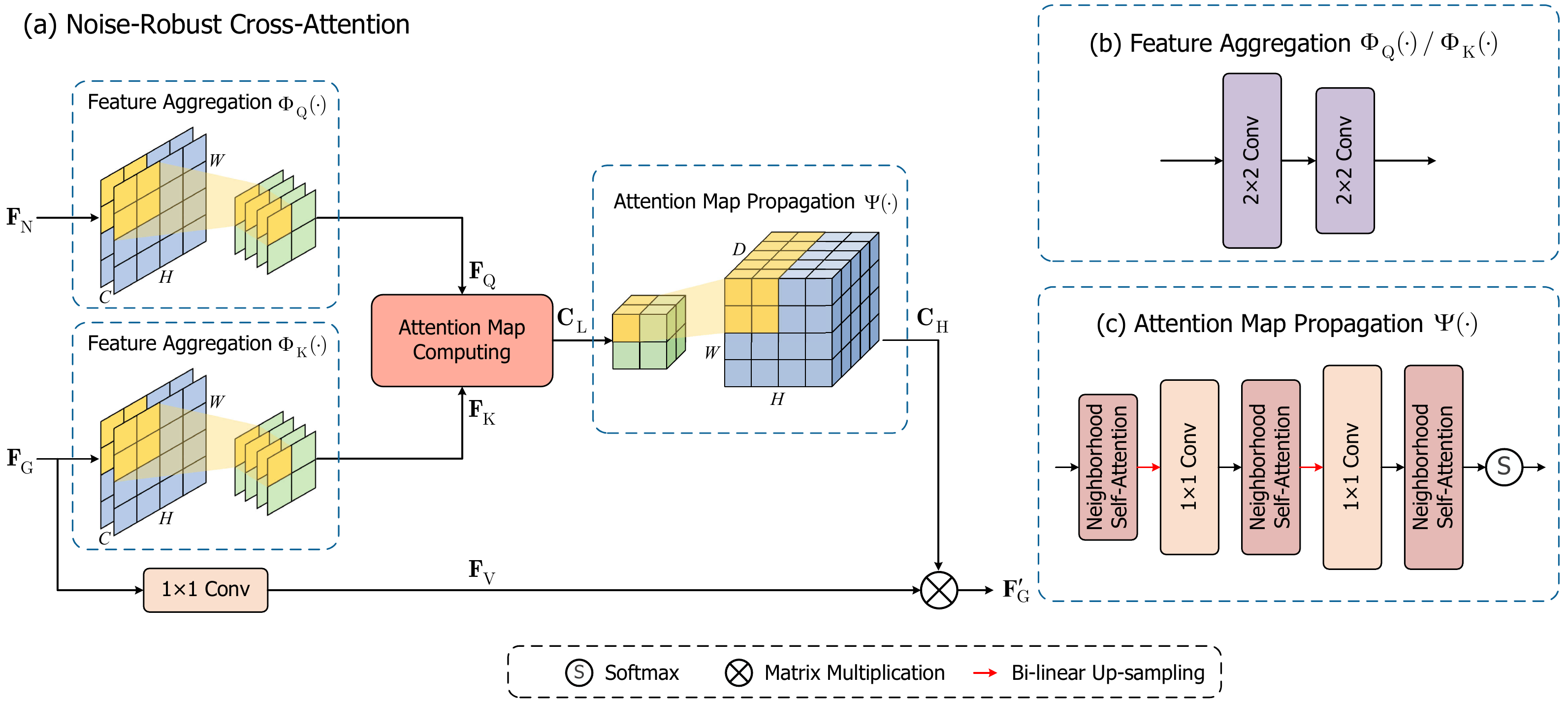}
	\caption{Illustration of (a) the noise-robust cross-attention (NRCA) module, which consists of (b) feature aggregation, coarse attention map computing, (c) attention map propagation, and aligned guidance feature generation.}
	\label{fig:NRCA} 
\end{figure}

We employ the cross-attention mechanism to aggregate the non-local information of $\mathbf{F}_\mathrm{G}$ to obtain the aligned guidance features $\mathbf{F}'_\mathrm{G} \in \mathbb{R}^{H\times W \times C}$. Specifically, the query features are generated from $\mathbf{F}_\mathrm{N}$, and the key/value features are generated from $\mathbf{F}_\mathrm{G}$. Since $\mathbf{I}_\mathrm{N}$ is seriously contaminated by noise, directly computing the pixel-level cross-attention map of $\mathbf{F}_\mathrm{N}$ and $\mathbf{F}_\mathrm{G}$ may have a large number of incorrect correspondences. To improve the robustness of cross-attention, the NRCA module adopts a four-step paradigm, including feature aggregation, coarse attention map computing, attention map propagation, and aligned guidance feature generation, as shown in Fig.~\ref{fig:NRCA} (a).

\textbf{Feature Aggregation.} Motivated by the observation that the correspondence between stereo images is locally consistent, and computing the attention map at a patch-level can alleviate the interference of noise, we aggregate $\mathbf{F}_\mathrm{N}$ and $\mathbf{F}_\mathrm{G}$ along spatial dimensions to generate the query $\mathbf{F}_\mathrm{Q}$ and key $\mathbf{F}_\mathrm{K}$ before the attention map computing. The process can be formulated as:
\begin{equation}
	\mathbf{F}_\mathrm{Q} = \Phi_\mathrm{Q}(\mathbf{F}_\mathrm{N}), \mathbf{F}_\mathrm{K} = \Phi_\mathrm{K}(\mathbf{F}_\mathrm{G}),
	\label{eq:feature_aggregator}
\end{equation}
where $\Phi_\mathrm{Q}(\cdot)$ and $\Phi_\mathrm{V}(\cdot)$ denote feature aggregators. As shown in Fig.~\ref{fig:NRCA} (b), the feature aggregator consists of two $2\times2$ convolution layers with stride $=2$, and the first layer extends the feature dimension from $C$ to $2C$, thus obtaining $\mathbf{F}_\mathrm{Q}$/$\mathbf{F}_\mathrm{K}$ of size  $\frac{H}{4}\times\frac{W}{4}\times{2C}$.

\textbf{Coarse Attention Map Computing.} We build the coarse attention map $\mathbf{C}_\mathrm{L} \in \mathbb{R}^{ \frac{H}{4}\times\frac{W}{4}\times\frac{D}{4}}$ within the range of max disparity, which can be formulated as
\begin{equation}
	\mathbf{C}_\mathrm{L}(h,w,d) = \langle\mathbf{F}_\mathrm{Q}(h,w), \mathbf{F}_\mathrm{K}(h,w-d)\rangle,
	\label{eq:coarse_attn}
\end{equation}
where $\langle\cdot,\cdot\rangle$ denotes the inner product, and $d$ is the disparity index.

\textbf{Attention Map Propagation.} The value of $\mathbf{C}_\mathrm{L}$ is simply computed by the inner product of single feature points, which lacks the ability to capture neighboring information. To address this limitation, we process $\mathbf{C}_\mathrm{L}$ through the network $\Psi(\cdot)$ to capture the piece-wise smoothness of neighboring regions, and expand the attention map to a pixel-level one $\mathbf{C}_\mathrm{H} \in \mathbb{R}^{H\times W \times D}$, formulated as
\begin{equation}
	\mathbf{C}_\mathrm{H} = \Psi(\mathbf{C}_\mathrm{L}).
	\label{eq:attn_refine}
\end{equation}
Each element in $\mathbf{C}_\mathrm{H}$ represents the attention score between coordinates $(h,w)$ in $\mathbf{F}_\mathrm{N}$ and $(h,w-d)$ in $\mathbf{F}_\mathrm{G}$. As show in Fig.~\ref{fig:NRCA} (c), $\Psi(\cdot)$ mainly consists of neighborhood self-attention (NSA)~\cite{Natten_CVPR23} layers, bi-linear up-sampling, $1\times1$ convolution layers, and the softmax function. Specifically, the NSA layer captures the self-similarity of stereo correspondence in a local area, aiming to remove the incorrect correspondences in the attention map, which can be formulated as
\begin{equation}
	\mathtt{NSA}(\mathtt{x}) = \mathtt{softmax}\left(\frac{\mathbf{Q}(\mathcal{\mathrm{x}})^{\mathrm{T}}\mathbf{K}(\mathcal{N(\mathrm{x})})}{\sqrt{d}}\right) \mathbf{V}(\mathcal{N(\mathrm{x})}),
	\label{eq:natten}
\end{equation}
where $\mathrm{x}$ is the coordinate index of the input feature map, $\mathcal{N(\mathrm{x})}$ is a local region around $\mathrm{x}$ with a window size $k$, $\mathbf{Q}$, $\mathbf{K}$, and $\mathbf{V}$ denote the projected query, key, and value, and $\sqrt{d}$ is the scaling parameter. The bi-linear up-sampling and $1\times1$ convolution layers are used to expand the spatial and disparity resolution of the input attention map, respectively. Finally, the softmax function is employed to normalize the attention map.

\textbf{Aligned Guidance Feature Generation.} We first use a $1\times1$ convolution layer to generate value $\mathbf{F}_\mathrm{V} \in \mathbb{R}^{H\times W\times\ C}$ from $\mathbf{F}_\mathrm{G}$, then perform a weighted summation of $\mathbf{F}_\mathrm{V}$ according to the pixel-wise attention map $\mathbf{C}_\mathrm{H}$ to get the aligned guidance features $\mathbf{F}'_\mathrm{G} \in \mathbb{R}^{H\times W \times C}$,
\begin{equation}
	\mathbf{F}'_\mathrm{G}(h,w) = \sum_{d=0}^{D} {\mathbf{C}_\mathrm{H}}(h,w,d) \mathbf{F}_\mathrm{V}(h,w-d).
	\label{eq:aggregation}
\end{equation}

\begin{figure}[t]
	\centering\includegraphics[scale=0.55]{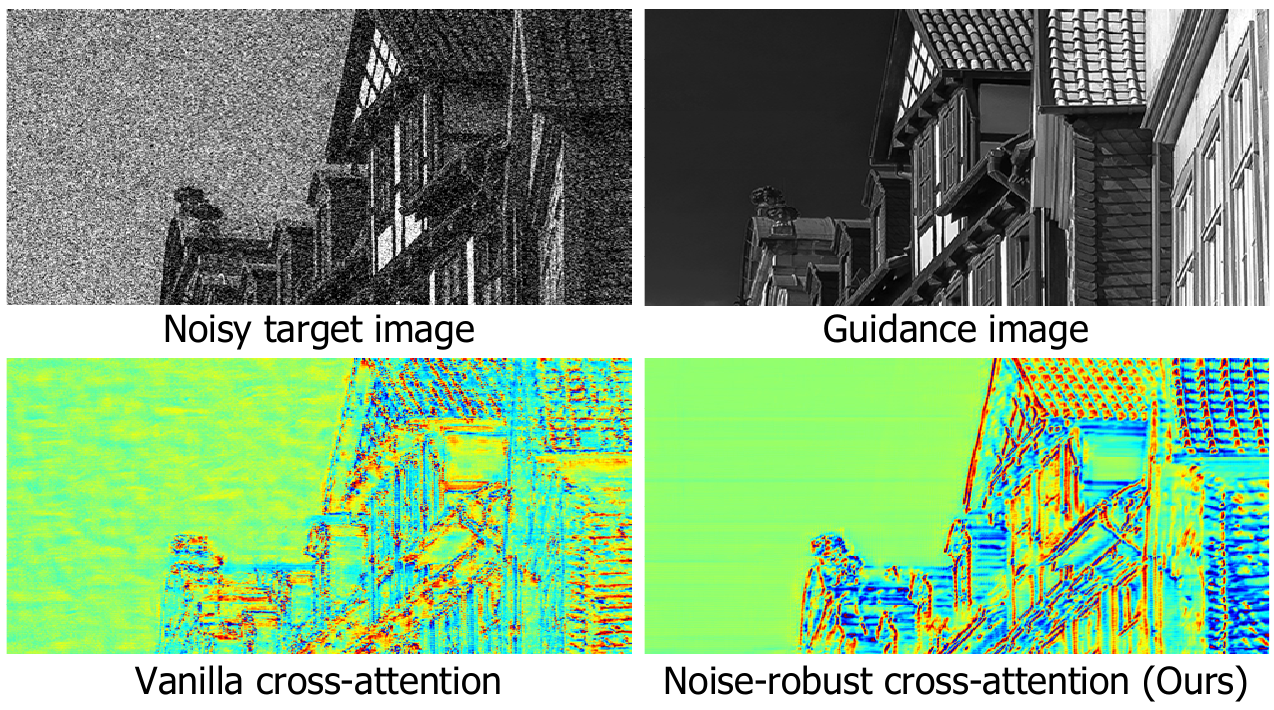}
	\caption{Visualization of aligned guidance feature maps. Compared to the vanilla cross-attention, the proposed noise-robust cross-attention can generate aligned guidance features with more salient structures under a high noise level.}
	\label{fig:NRCA_feature} 
\end{figure}

\textbf{Discussions.} It should be noted that there is a clear difference between the proposed NRCA module and the cross-attention mechanism used in stereo image restoration~\cite{PASSRnet_CVPR19, SIR-Former_ACMMM22, NAFSSR_CVPRW22, ACLRNet_TPAMI24}. Existing approaches typically employ the vanilla cross-attention to directly compute a pixel-level attention map. On the contrary, the feature aggregation and attention map propagation enable our NRCA module to capture precise correspondence between the noisy target and guidance image, thus obtaining high-quality aligned guidance features. As shown in Fig.~\ref{fig:NRCA_feature}, compared to the vanilla cross-attention, our NRCA module can effectively avoid the interference of noise, and generates aligned guidance features with more salient structures and details. More ablations in Section~\ref{subsec:ablation} further validate the effectiveness of the NRCA module in denoising results.

\subsection{Spatially Variant Feature Fusion (SVFF)}

\begin{figure}[t]
	\centering\includegraphics[scale=0.22]{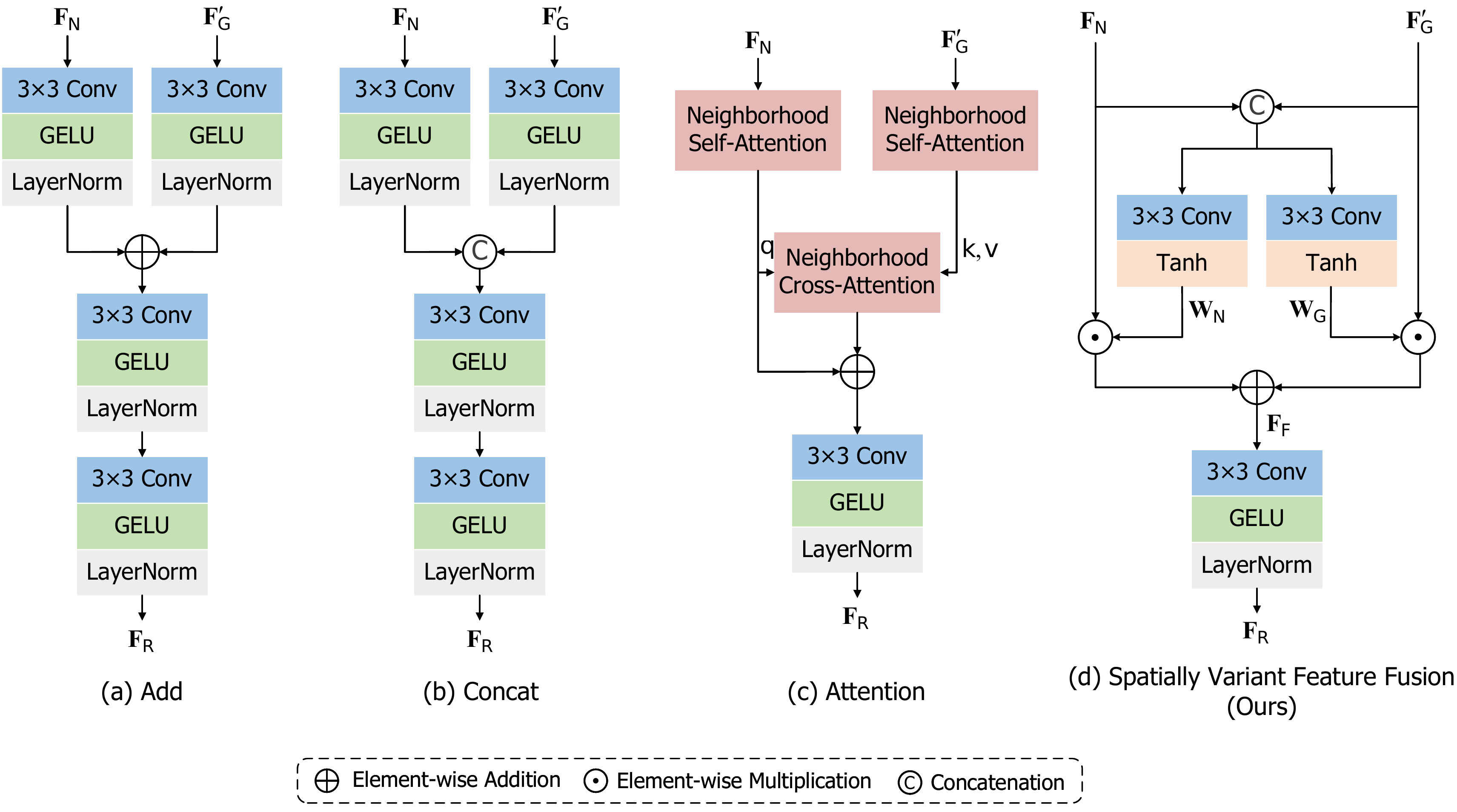}
	\caption{Comparison of different feature fusion strategies. (a) Add. (b) Concat. (c) Attention. (d) Our proposed spatially variant feature fusion (SVFF) module.}
	\label{fig:SVFF} 
\end{figure}

After obtaining the aligned guidance features $\mathbf{F}'_\mathrm{G}$, we introduce the SVFF module to fuse $\mathbf{F}_\mathrm{N}$ and $\mathbf{F}'_\mathrm{G}$ by predicting spatially variant weights according to the content of feature. By integrating the NRCA module and the SVFF module into a unified transformer architecture, this fusion strategy can further enhance the fine structures and suppress the harmful artifacts of input features in a simple but effective way.

As illustrated in Fig.~\ref{fig:SVFF}(d), we concatenate $\mathbf{F}_\mathrm{N}$ and $\mathbf{F}'_\mathrm{G}$ along the feature dimension, and use two $3\times3$ convolution layers with $\mathtt{tanh}$ as activation function to generate the spatially variant weights $\mathbf{W}_\mathrm{N}, \mathbf{W}_\mathrm{G} \in \mathbb{R}^{H\times W \times C}$. We then perform the element-wise multiplication and addition to get the fused features $\mathbf{F}_\mathrm{F} \in \mathbb{R}^{H\times W \times C}$,
\begin{equation}
	\mathbf{F}_\mathrm{F} = \mathbf{F}_\mathrm{N} \odot \mathbf{W}_\mathrm{N} + \mathbf{F}'_\mathrm{G} \odot \mathbf{W}_\mathrm{G},
	\label{eq:svff}
\end{equation}
where $\odot$ denotes the element-wise multiplication. Finally, we use a $3\times3$ convolution layer to refine $\mathbf{F}_\mathrm{F}$. The refined features $\mathbf{F}_\mathrm{R} \in \mathbb{R}^{H\times W \times C}$ will be used as the input of the feed-forward network of the transformer block.

\begin{figure}[t]
	\centering\includegraphics[scale=1]{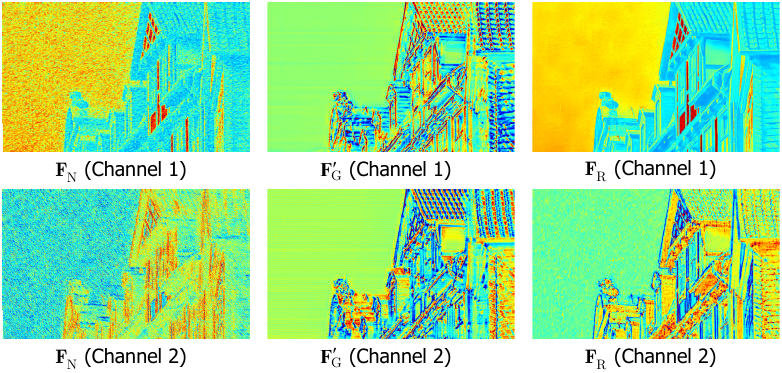}
	\caption{Visualization of the input and output feature maps of the spatially variant feature fusion module.}
	\label{fig:SVFF_feature} 
\end{figure}

\textbf{Discussion.} To figure out the effectiveness of our proposed SVFF module, we compare it with several different feature fusion strategies that widely used in previous works~\cite{DarkVisionNet_AAAI22, Fusion_IF22, SwinFusion_JAS22}: Add (Fig.~\ref{fig:SVFF} (a)), Concat (Fig.~\ref{fig:SVFF} (b)), and Attention (Fig.~\ref{fig:SVFF} (c)). The denoising performance of different configurations of feature fusion are shown in Table~\ref{tab:ablation} (e), demonstrating that our SVFF module outperforms other comparative feature fusion strategies with similar computational cost. We display the input and output feature maps of the SVFF module in Fig.~\ref{fig:SVFF_feature}. Since the spatially variant weights are channel independent, the SVFF module can adaptively select useful information for denoising per channel, including both low frequency and high frequency components.

\section{Experiments}
\label{sec:exp}

\subsection{Datasets and Metrics}
\textbf{Synthetic Noise Datasets.} We evaluate our method on three synthetic noise datasets, \ie the PittsStereo-RGBNIR dataset~\cite{Pitts_CVPR18}, the Flickr1024 dataset~\cite{Flickr_ICCVW19}, and the KITTI Stereo 2015 dataset~\cite{Kitti_CVPR15}. The PittsStereo-RGBNIR dataset contains 40,000 RGB-NIR stereo image pairs for training and 2,000 for testing. The latter two datasets are RGB-RGB stereo image pairs. Following~\cite{SANet_CVPR23}, we use the G, B, and R channels of the right-view images as guidance to denoise the R, G, and B channels of the left-view noisy images, respectively. Since the network only processes one channel at a time, it is guaranteed that the target image and the guidance image are captured in different spectral bands. Specifically, the Flickr1024 dataset includes 800 image pairs for training and 112 for testing, and the KITTI Stereo 2015 dataset contains 400 image pairs for training and 400 for testing.

\textbf{Real-world Noise Dataset.} To further explore the performance under real noise, we evaluate our method on the Dark Flash Stereo dataset~\cite{SDF_ICCP19}, which is captured through a stereo system with a RGB camera and a NIR-G-NUV (near ultraviolet) camera. We select 21 scenes for training and 11 scenes for testing. Each scene contains 4 stereo image pairs obtained under different exposure times, and a long-exposure ground-truth image of the RGB camera. The different exposure times indicate different noise levels. We use the NIR channel as guidance to denoise the R, G, and B channels, respectively.

\textbf{Metrics.} We employ peak signal to noise ratio (PSNR), structural similarity (SSIM)~\cite{SSIM_TIP04}, and learned perceptual image patch similarity (LPIPS)~\cite{Perceptual_ECCV16} to quantitatively measure the denoising performance of different methods. Higher PSNR and SSIM, lower LPIPS indicate better denoising performance.

\subsection{Implementation Details}
To demonstrate the effectiveness and potential of our method, we present two versions of our network: SGDFormer with the number of transformer block $L=1$, and SGDFormer{$^\dagger$}  with $L=3$.

The network is implemented with the Pytorch framework. The maximal disparity $D$ and the channel number $C$ are set to $128$ and $32$, respectively. Balancing denoising performance and computation cost, the window size $k$ of neighborhood self-attention in the attention map propagation is set to $5$. We adopt the AdamW optimizer~\cite{AdamW} with $\beta_{1}=0.9, \beta_{2}=0.99$, and use the cosine decay strategy~\cite{Cosine} to decrease the learning rate from $5\times 10^{-4}$ to $1\times 10^{-6}.$ The batch size is set to $8$, and the patch size is set to $128\times400$.

For experiments on the synthetic noise datasets, we add Poisson-Gaussian noise~\cite{PGNoise_TIP08} to the clean target images to simulate noisy samples. During training, we randomly sample the Poisson noise parameter $\alpha\in[0, 0.02]$ and Gaussian noise parameter $\sigma\in[0, 0.2]$. The number of total training iterations is 200,000. For real-world noise evaluation on the Dark Flash Stereo dataset, we fine-tune the model trained on the Flickr1024 dataset for 20,000 iterations. The comparison approaches adopt the same training strategy for fairness. All the experiments are conducted on an NVIDIA Geforce RTX 4090 GPU.

\subsection{Evaluation on Synthetic Noise Datasets}

\begin{figure*}[t]
	\centering\includegraphics[scale=0.68]{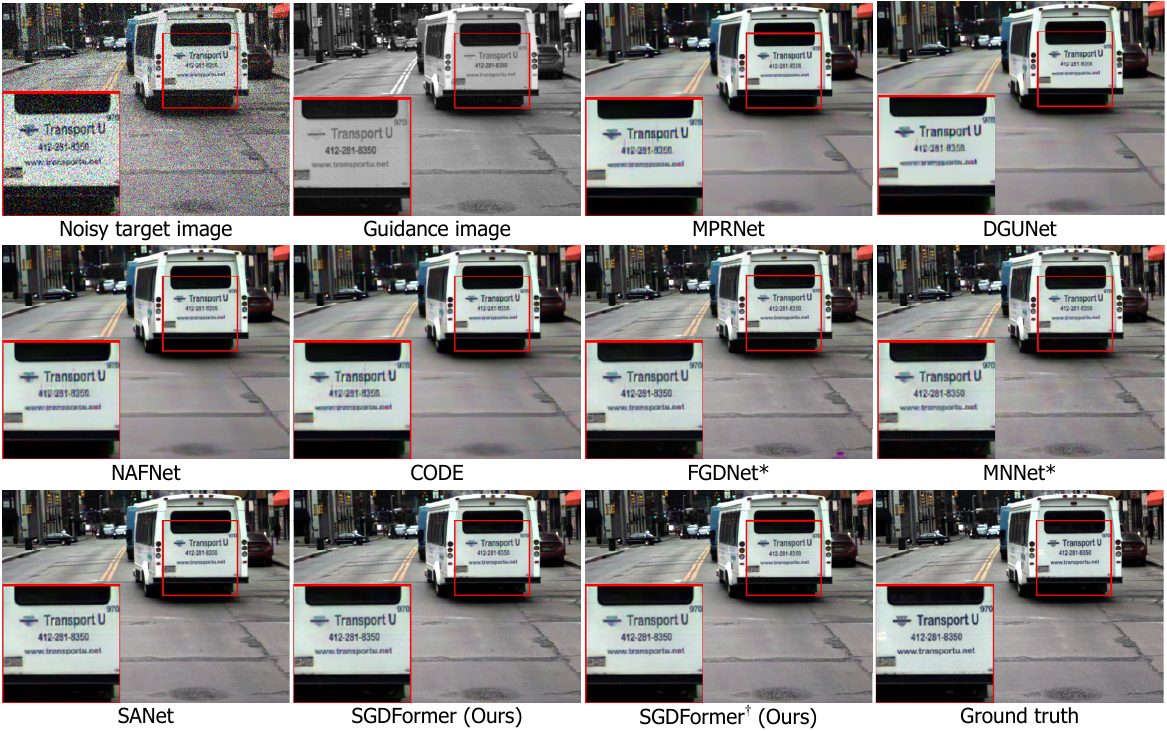}
	\caption{Qualitative comparison of different denoising methods on the PittsStereo-RGBNIR dataset under Poisson-Gaussian noise ($\alpha=0.02, \sigma=0.2$).}
	\label{fig:Res_DMC_PG}
\end{figure*}

\begin{figure*}[t]
	\centering\includegraphics[scale=0.68]{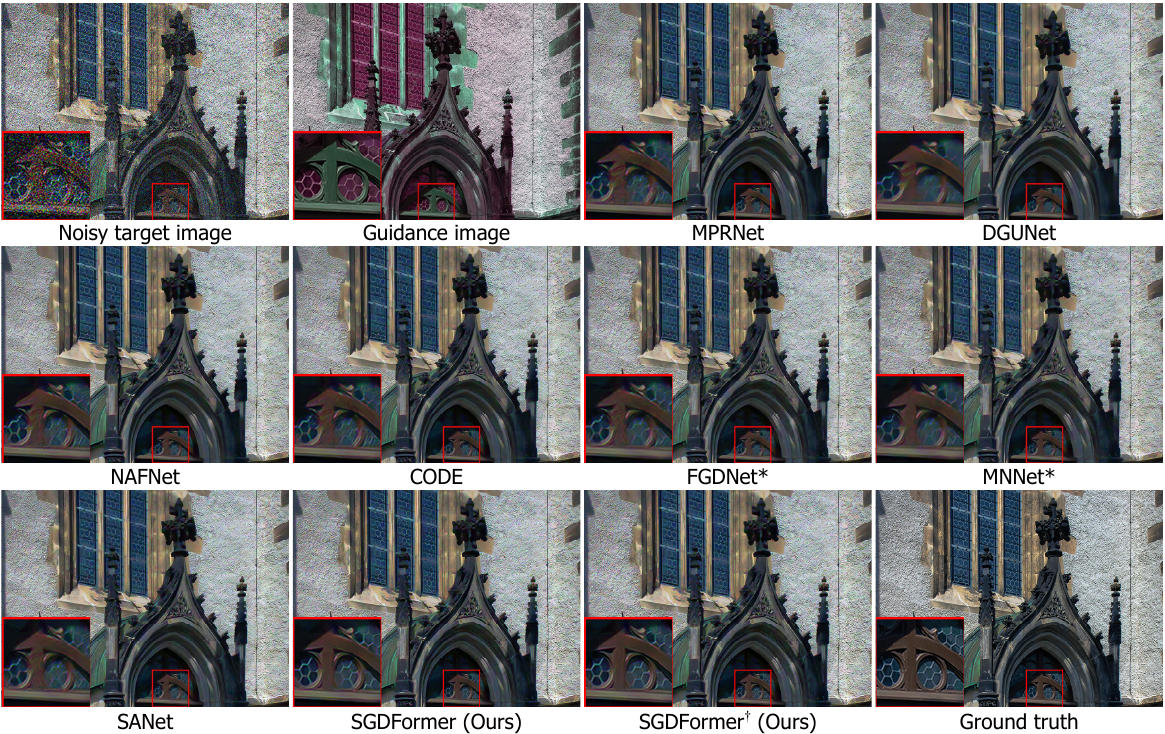}
	\caption{Qualitative comparison of different denoising methods on the Flickr1024 dataset under Gaussian noise ($\sigma=0.2$).}
	\label{fig:Res_Flickr_G}
\end{figure*}

\begin{figure*}[t]
	\centering\includegraphics[scale=0.68]{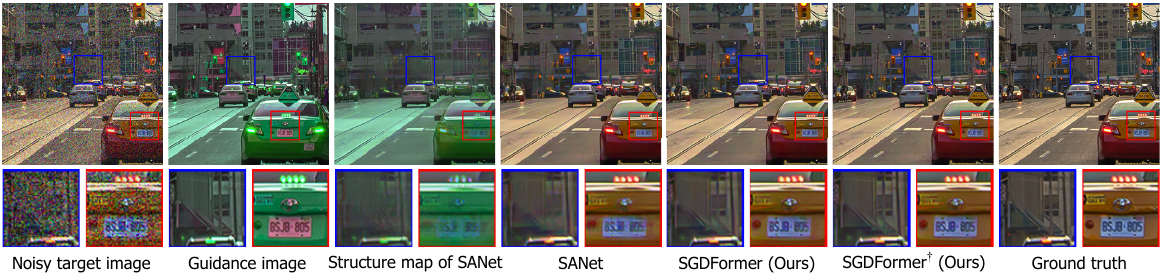}
	\caption{Qualitative comparison of our method (SGDFormer and SGDFormer{$^\dagger$}) with the two-step approach SANet on the Flickr1024 dataset under Poisson-Gaussian noise ($\alpha=0.02, \sigma=0.2$).}
	\label{fig:Compare_with_SANet}
\end{figure*}

\begin{figure*}[t]
	\centering\includegraphics[scale=0.68]{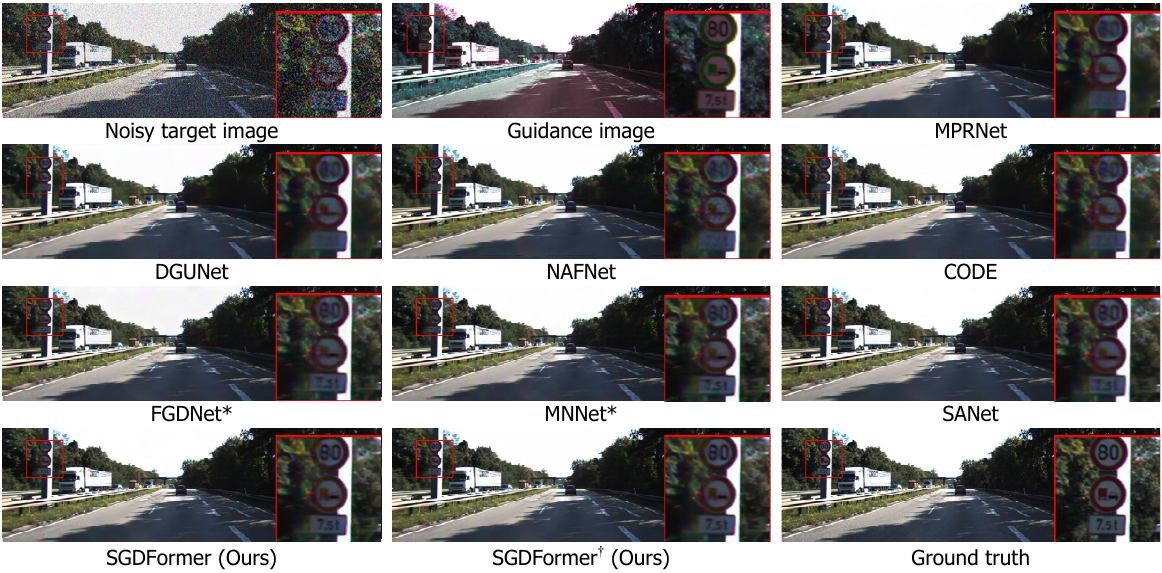}
	\caption{Qualitative comparison of different denoising methods on the KiTTI Stereo 2015 dataset under Poisson-Gaussian noise ($\alpha=0.02, \sigma=0.2$).}
	\label{fig:Res_Kitti_PG}
\end{figure*}

\begin{table}
	\caption{Quantitative comparison (PSNR/SSIM/LPIPS) on the PittsStereo-RGBNIR dataset under Gaussian noise ($\sigma=0.2$) and Poisson-Gaussian noise ($\alpha=0.02,\sigma=0.2$). The best two results are highlighted in {\color{red}{red}} and {\color{blue}{blue}}, respectively.}
	\renewcommand{\tabcolsep}{5pt}{}
	\renewcommand\arraystretch{1.2}
	\centering
	\resizebox{\linewidth}{!}
	{
		\begin{tabular}{c|ccc|ccc}
			\hline
			\multirow{2.1}{*}{Method} & \multicolumn{3}{c|}{$\sigma=0.2$} & \multicolumn{3}{c}{$\alpha=0.02$, $\sigma=0.2$} \\
			\cline{2-7}
			& PSNR $\uparrow$ & SSIM $\uparrow$ & LPIPS $\downarrow$ & PSNR $\uparrow$ & SSIM $\uparrow$ & LPIPS $\downarrow$ \\
			\hline
			MPRNet~\cite{MPRNet_CVPR21} & 27.74 & 0.8416 & 0.3861 & 27.34 & 0.8349 & 0.3967 \\
			DGUNet~\cite{DGUNet_CVPR22} & 27.79 & 0.8390 & 0.3825 & 27.39 & 0.8336 & 0.3913 \\
			Uformer~\cite{Uformer_CVPR22} & 27.67 & 0.8383 & 0.3985 & 27.26 & 0.8319 & 0.4023 \\
			NAFNet~\cite{NAFNet_ECCV22} & 27.94 & 0.8471 & 0.3238 & 27.57 & 0.8415 & 0.3364 \\
			CODE~\cite{CODE_CVPR23} & 27.97 & 0.8466 & 0.3687 & 27.57 & 0.8411 & 0.3739 \\
			\hline
			FGDNet*~\cite{FGDNet_TMM22} & 28.49 & 0.8695 & 0.2920 & 28.10 & 0.8647 & 0.2995 \\
			MNNet*~\cite{MNNet_IF22} & 29.28 & 0.8749 & 0.2606 & 28.85 & 0.8710 & 0.2689 \\
			SANet~\cite{SANet_CVPR23} & 29.32 & 0.8761 & 0.2565 & 28.98 & 0.8726 & 0.2606 \\
			\hline
			SGDFormer (Ours) & {\color{blue}{29.51}} & {\color{blue}{0.8782}} & {\color{blue}{0.2461}} & {\color{blue}{29.17}} & {\color{blue}{0.8745}} & {\color{blue}{0.2520}} \\
			SGDFormer{$^\dagger$} (Ours) & {\color{red}{29.72}} & {\color{red}{0.8823}} & {\color{red}{0.2370}} & {\color{red}{29.39}} & {\color{red}{0.8786}} & {\color{red}{0.2425}} \\
			\hline
		\end{tabular}
	}
	\label{tab:comparison_DMC}
\end{table}

\begin{table}
	\caption{Quantitative comparison (PSNR/SSIM/LPIPS) on the Flickr1024 dataset under Gaussian noise ($\sigma=0.2$) and Poisson-Gaussian noise ($\alpha=0.02,\sigma=0.2$). The best two results are highlighted in {\color{red}{red}} and {\color{blue}{blue}}, respectively.}
	\renewcommand{\tabcolsep}{5pt}{}
	\renewcommand\arraystretch{1.2}
	\centering
	\resizebox{\linewidth}{!}
	{
		\begin{tabular}{c|ccc|ccc}
			\hline
			\multirow{2.1}{*}{Method} & \multicolumn{3}{c|}{$\sigma=0.2$} & \multicolumn{3}{c}{$\alpha=0.02$, $\sigma=0.2$} \\
			\cline{2-7}
			& PSNR $\uparrow$ & SSIM $\uparrow$ & LPIPS $\downarrow$ & PSNR $\uparrow$ & SSIM $\uparrow$ & LPIPS $\downarrow$ \\
			\hline
			MPRNet~\cite{MPRNet_CVPR21} & 25.33 & 0.8370 & 0.2968 & 24.95 & 0.8287 & 0.3058 \\
			DGUNet~\cite{DGUNet_CVPR22} & 25.32 & 0.8294 & 0.2953 & 24.96 & 0.8229 & 0.3043 \\
			Uformer~\cite{Uformer_CVPR22} & 25.15 & 0.8302 & 0.3052 & 24.76 & 0.8225 & 0.3102 \\
			NAFNet~\cite{NAFNet_ECCV22} & 25.56 & 0.8427 & 0.2844 & 25.20 & 0.8356 & 0.2936 \\
			CODE~\cite{CODE_CVPR23} & 25.47 & 0.8397 & 0.2884 & 25.09 & 0.8326 & 0.2945 \\
			\hline
			FGDNet*~\cite{FGDNet_TMM22} & 25.23 & 0.8337 & 0.2846  & 24.83 & 0.8264 & 0.2923 \\
			MNNet*~\cite{MNNet_IF22} & 25.66 & 0.8456 & 0.2799  & 25.30 & 0.8367 & 0.2867 \\
			SANet~\cite{SANet_CVPR23} & 25.67 & 0.8477 & 0.2685 & 25.30 & 0.8411 & 0.2736 \\
			\hline
			SGDFormer (Ours) & {\color{blue}{26.01}} & {\color{blue}{0.8589}} & {\color{blue}{0.2455}} & {\color{blue}{25.65}} & {\color{blue}{0.8525}} & {\color{blue}{0.2521}} \\
			SGDFormer{$^\dagger$} (Ours) & {\color{red}{26.15}} & {\color{red}{0.8643}} & {\color{red}{0.2437}} & {\color{red}{25.79}} & {\color{red}{0.8579}} & {\color{red}{0.2501}} \\
			\hline
		\end{tabular}
	}
	\label{tab:comparison_Flickr}
\end{table}

\begin{table}
	\caption{Quantitative comparison (PSNR/SSIM/LPIPS) on the KITTI Stereo 2015 dataset under Gaussian noise ($\sigma=0.2$) and Poisson-Gaussian noise ($\alpha=0.02,\sigma=0.2$). The best two results are highlighted in {\color{red}{red}} and {\color{blue}{blue}}, respectively.}
	\renewcommand{\tabcolsep}{5pt}{}
	\renewcommand\arraystretch{1.2}
	\centering
	\resizebox{\linewidth}{!}
	{
		\begin{tabular}{c|ccc|ccc}
			\hline
			\multirow{2.1}{*}{Method} & \multicolumn{3}{c|}{$\sigma=0.2$} & \multicolumn{3}{c}{$\alpha=0.02$, $\sigma=0.2$} \\
			\cline{2-7}
			& PSNR $\uparrow$ & SSIM $\uparrow$ & LPIPS $\downarrow$ & PSNR $\uparrow$ & SSIM $\uparrow$ & LPIPS $\downarrow$ \\
			\hline
			MPRNet~\cite{MPRNet_CVPR21} & 26.59 & 0.8580 & 0.3308 & 26.18 & 0.8509 & 0.3407 \\
			DGUNet~\cite{DGUNet_CVPR22} & 26.55 & 0.8577 & 0.3287 & 26.12 & 0.8505 & 0.3415 \\
			Uformer~\cite{Uformer_CVPR22} & 26.68 & 0.8609 & 0.3270 & 26.27 & 0.8546 & 0.3313 \\
			NAFNet~\cite{NAFNet_ECCV22} & 27.02 & 0.8665 & 0.2986 & 26.60 & 0.8602 & 0.3075 \\
			CODE~\cite{CODE_CVPR23} & 27.53 & 0.8750 & 0.2867 & 27.08 & 0.8688 & 0.2951 \\
			\hline
			FGDNet*~\cite{FGDNet_TMM22} & 27.57 & 0.8834 & 0.2505 & 27.17 & 0.8805 & 0.2569 \\
			MNNet*~\cite{MNNet_IF22} & 27.79 & 0.8856 & 0.2429 & 27.36 & 0.8820 & 0.2527 \\
			SANet~\cite{SANet_CVPR23} & 27.88 & 0.8899 & 0.2439 & 27.47 & 0.8851 & 0.2513 \\
			\hline
			SGDFormer (Ours) & {\color{blue}{28.21}} & {\color{blue}{0.8985}} & {\color{blue}{0.2222}} & {\color{blue}{27.80}} & {\color{blue}{0.8939}} & {\color{blue}{0.2284}} \\
			SGDFormer{$^\dagger$} (Ours) & {\color{red}{28.39}} & {\color{red}{0.9018}} & {\color{red}{0.2151}} & {\color{red}{27.98}} & {\color{red}{0.8973}} & {\color{red}{0.2213}} \\
			\hline
		\end{tabular}
	}
	\label{tab:comparison_Kitti}
\end{table}

We compare our SGDFormer and SGDFormer{$^\dagger$} with eight image denoising approaches, including MPRNet~\cite{MPRNet_CVPR21}, DGUNet~\cite{DGUNet_CVPR22}, Uformer~\cite{Uformer_CVPR22}, NAFNet~\cite{NAFNet_ECCV22}, CODE~\cite{CODE_CVPR23}, FGDNet~\cite{FGDNet_TMM22}, MNNet~\cite{MNNet_IF22}, and SANet~\cite{SANet_CVPR23}. The first five ones are single image denoising approaches, and the rest are guided denoising ones. Since FGDNet and MNNet assume that the guidance image and target image are spatially-aligned, we adopt the structure map generated by SANet as the guidance, denoted as FGDNet* and MNNet*, respectively. We evaluate the denoising performance of different methods under Gaussian noise ($\sigma=0.2$) and Poisson-Gaussian noise ($\alpha=0.02, \sigma=0.2$).

Table~\ref{tab:comparison_DMC} shows the quantitative results on the PittsStereo-RGBNIR dataset, from which we can observe that our SGDFormer and SGDFormer{$^\dagger$} outperform all other approaches by a large margin. Compared with the previous state-of-the-art approach SANet, our SGDFormer and SGDFormer{$^\dagger$} increase PSNR by 0.19dB and 0.41dB under Poisson-Gaussian noise ($\alpha=0.02, \sigma=0.2$). Fig.~\ref{fig:Res_DMC_PG} displays the visual comparison of different methods. Although single image denoising approaches can suppress noises, they cannot preserve structures and textures well. Among all guided denoising methods, our network can produce denoised images with more salient structures.

We list the quantitative results on the Flickr1024 dataset in Table~\ref{tab:comparison_Flickr}. Our SGDFormer and SGDFormer{$^\dagger$} achieve the highest denoising accuracy under different noise levels. Compared with guided denoising approaches FGDNet*, MNNet*, and SANet, our SGDFormer achieves 0.78 dB, 0.35 dB, and 0.34 dB PSNR gains under Gaussian noise ($\sigma=0.2$) and 0.82 dB, 0.35 dB, and 0.35 dB PSNR gains under Poisson-Gaussian noise ($\alpha=0.02, \sigma=0.2$). This indicates that our method has a better utility of the guidance image. The qualitative results in Fig.~\ref{fig:Res_Flickr_G} show that our method restores more fine-scale textures and has fewer artifacts than other competitors. Furthermore, Fig.~\ref{fig:Compare_with_SANet} presents the visual results of SANet and our method on the Flickr1024 dataset. It is observed that the structure map of SANet cannot preserve the structures and details of the original guidance image, leading to ghosting and artifacts of the denoised image. Benefiting from the one-stage end-to-end architecture, our method can directly transfer fine structures from the guidance image to the target image without producing artifacts.

The quantitative results on the Kitti Stereo 2015 dataset are presented in Table~\ref{tab:comparison_Kitti}, where our SGDFormer and SGDFormer{$^\dagger$} consistently maintain superior performance, transcending all other methods. As shown in Fig.~\ref{fig:Res_Kitti_PG}, the single image denoising approaches cannot restore the text and icons on the road sign. Compared with other guided denoising approaches, our method can capture more accurate stereo correspondence, thus recovering more salient structures.

\subsection{Evaluation on the Real-world Noise Dataset}

\begin{figure*}[t]
	\centering\includegraphics[scale=0.68]{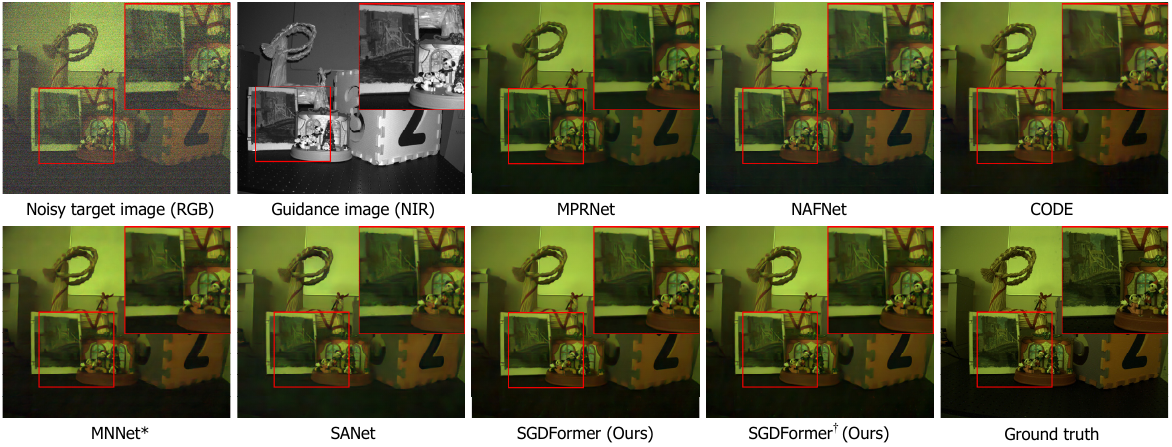}
	\caption{Qualitative comparison of different denoising methods on the Dark Flash Stereo dataset.}
	\label{fig:Res_DFS}
\end{figure*}

\begin{table}
	\caption{Quantitative comparison (PSNR/SSIM/LPIPS) on the Dark Flash Stereo dataset. The best two results are highlighted in {\color{red}{red}} and {\color{blue}{blue}}, respectively.}
	\renewcommand{\tabcolsep}{25pt}{}
	\renewcommand\arraystretch{1.2}
	\centering
	\resizebox{\linewidth}{!}
	{
		\begin{tabular}{c|ccc}
			\hline
			Method & PSNR $\uparrow$ & SSIM $\uparrow$ & LPIPS $\downarrow$ \\
			\hline
			MPRNet~\cite{MPRNet_CVPR21} & 31.35 & 0.9367 & 0.2120\\
			DGUNet~\cite{DGUNet_CVPR22} & 31.79 & 0.9152 & 0.2020 \\
			Uformer~\cite{Uformer_CVPR22} & 31.24 & 0.8990 & 0.2030 \\
			NAFNet~\cite{NAFNet_ECCV22} & 32.10 & 0.9286 & 0.2043 \\
			CODE~\cite{CODE_CVPR23} & 31.89 & 0.9475 & 0.2115 \\
			\hline
			FGDNet*~\cite{FGDNet_TMM22} & 31.90 & 0.9381 & 0.2117 \\
			MNNet*~\cite{MNNet_IF22} & 32.62 & 0.9499 & 0.2122 \\
			SANet~\cite{SANet_CVPR23} & 31.92 & 0.9418 & 0.2106\\
			\hline
			SGDFormer (Ours) & {\color{blue}{33.38}} & {\color{blue}{0.9604}} & {\color{blue}{0.1939}} \\
			SGDFormer{$^\dagger$} (Ours) & {\color{red}{33.58}} & {\color{red}{0.9632}} & {\color{red}{0.1905}} \\
			\hline
		\end{tabular}
	}
	\label{tab:comparison_DFS}
\end{table}

To demonstrate the generalizability of our method under real-world noise, we evaluate the denoising performance on the Dark Flash Stereo dataset, and list the qualitative results in Table~\ref{tab:comparison_DFS}. Our SGDFormer and SGDFormer{$^\dagger$} surpass all competitors in terms of PSNR, SSIM, and LPIPS. As illustrated in Fig.~\ref{fig:Res_DFS}, the results produced by our method can effectively remove noise with fewer artifacts when compared with other approaches.

\subsection{Computation Cost}

\begin{figure*}[t]
	\centering\includegraphics[scale=0.28]{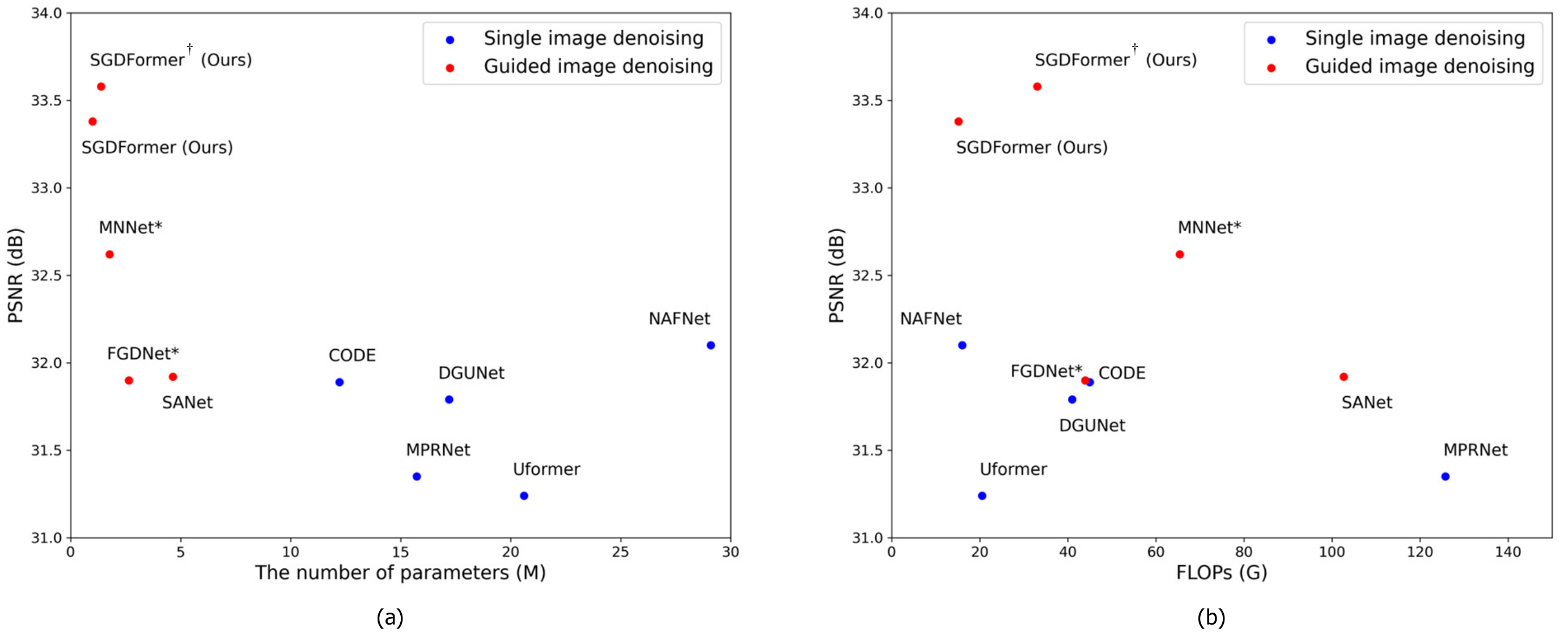}
	\caption{Denoising performance and computation cost comparisons on the Dark Flash Stereo dataset. The computation cost is measured with inputs of resolution $256\times256$. (a) PSNR vs. The number of parameters. (b) PSNR vs. FLOPs.}
	\label{fig:Computation_cost}
\end{figure*}

To further validate the superiority of our method, we presents the denoising performance and computation cost comparisons on the Dark Flash Stereo Dataset in Fig.~\ref{fig:Computation_cost}. We compute the number of parameters and floating point operations (FLOPs) of different networks. It can be observed that guided image denoising approaches generally have less parameters than the single denoising ones. Compared with other networks, our SGDFormer and SGDFormer{$^\dagger$} have the minimum parameters while obtaining the best denoising accuracy. Besides, our method has relatively small FLOPs, indicating a good balance between inference speed and denoising performance.

\subsection{Ablation Study}
\label{subsec:ablation}

\begin{table}
	\caption{Ablation study of SGDFormer.}
	\renewcommand{\tabcolsep}{5pt}{}
	\renewcommand\arraystretch{1}
	\centering
	\resizebox{\linewidth}{!}
	{
		\begin{tabular}{c|ccc|c}
			\multicolumn{5}{c}{(a) The influence of guidance image.} \\
			\hline
			Setting & PSNR $\uparrow$ & SSIM $\uparrow$ & LPIPS $\downarrow$ & Parameters (M) \\
			\hline
			w/o guidance image & 24.96 & 0.8260 & 0.3068 & 0.99\\
			w/ guidance image & 25.65 & 0.8525 & 0.2521 & 0.99\\
			\hline
			\multicolumn{5}{c}{ } \\
			\multicolumn{5}{c}{(b) The blocks of image encoder.} \\
			\hline
			Setting & PSNR $\uparrow$ & SSIM $\uparrow$ & LPIPS $\downarrow$ & Parameters (M) \\
			\hline
			ResBlock & 25.67 & 0.8540 & 0.2510 & 1.91 \\
			NAFBlock & 25.65 & 0.8525 & 0.2521 & 0.99 \\
			\hline
			\multicolumn{5}{c}{ } \\
			\multicolumn{5}{c}{(c) The components of transformer block.} \\
			\hline
			Setting & PSNR $\uparrow$ & SSIM $\uparrow$ & LPIPS $\downarrow$ & Parameters (M) \\
			\hline
			Vanilla transformer & 25.20 & 0.8357 & 0.2756 & 0.79 \\
			w/ NRCA & 25.38 & 0.8451 & 0.2568 & 0.94 \\
			w/ SVFF & 25.50 & 0.8455 & 0.2650 & 0.84 \\
			Full model & 25.65 & 0.8525 & 0.2521 & 0.99 \\
			\hline
			\multicolumn{5}{c}{ } \\
			\multicolumn{5}{c}{(d) Different attention map propagation strategies.} \\
			\hline
			Setting & PSNR $\uparrow$ & SSIM $\uparrow$ & LPIPS $\downarrow$ & Parameters (M) \\
			\hline
			Conv & 25.59 & 0.8504 & 0.2586 & 0.99 \\
			Neighborhood self-attention & 25.65 & 0.8525 & 0.2521 & 0.99 \\
			\hline
			\multicolumn{5}{c}{ } \\
			\multicolumn{5}{c}{(e) Different feature fusion strategies.} \\
			\hline
			Setting & PSNR $\uparrow$ & SSIM $\uparrow$ & LPIPS $\downarrow$ & Parameters (M) \\
			\hline
			Add & 25.54 & 0.8506 & 0.2533 & 0.98 \\
			Concat & 25.56 & 0.8506 & 0.2536 & 0.99 \\
			Attention & 25.57 & 0.8506 & 0.2548 & 0.96 \\
			SVFF & 25.65 & 0.8525 & 0.2521 & 0.99 \\
			\hline
			\multicolumn{5}{c}{ } \\
			\multicolumn{5}{c}{(f) The window size of neighborhood self-attention in the attention map propagation.} \\
			\hline
			Window size & PSNR $\uparrow$ & SSIM $\uparrow$ & LPIPS $\downarrow$ & Parameters (M)\\
			\hline
			3 & 25.61 & 0.8504 & 0.2576 & 0.99 \\
			5 & 25.65 & 0.8525 & 0.2521 & 0.99 \\
			7 & 25.65 & 0.8527 & 0.2509 & 0.99 \\
			\hline
			\multicolumn{5}{c}{ } \\
			\multicolumn{5}{c}{(g) The number of transformer blocks.} \\
			\hline
			\# Blocks & PSNR $\uparrow$ & SSIM $\uparrow$ & LPIPS $\downarrow$ & Parameters (M)\\
			\hline
			1 & 25.65 & 0.8525 & 0.2521 & 0.99 \\
			2 & 25.72 & 0.8560 & 0.2502 & 1.18 \\
			3 & 25.79 & 0.8579 & 0.2501 & 1.38 \\
			4 & 25.82 & 0.8592 & 0.2463 & 1.58 \\
			\hline
		\end{tabular}
	}
	\label{tab:ablation}
\end{table}

\begin{figure*}[t]
	\centering\includegraphics[scale=0.68]{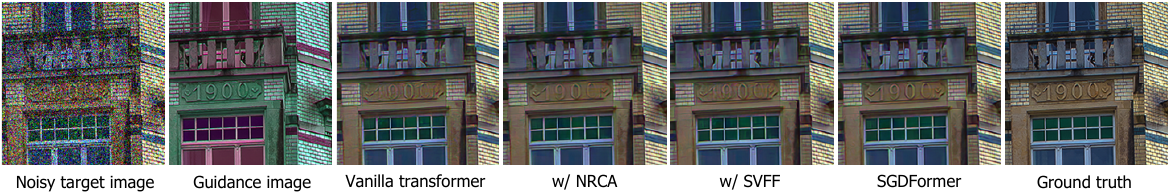}
	\caption{Qualitative comparison of the ablation study on the components of transformer block.}
	\label{fig:Compare_ablation}
\end{figure*}

We conduct the ablation study on the Flickr1024 dataset under Poisson-Gaussian noise ($\alpha=0.02, \sigma=0.2$).

\textbf{The influence of guidance image.} To demonstrate whether the guidance image is beneficial for denoising, we take the noisy target image as the guidance input of the network, without changing the network architecture. As shown in Table~\ref{tab:ablation} (a), in this case, SGDFormer will become a single image denoising network, leading to a significant decrease of denoising performance.

\textbf{The blocks of image encoder.} To demonstrate that the performance of SGDFormer does not depend on the specific network design of image encoder, we replace the NAFBlock with ResBlock~\cite{ResNet_CVPR16}, and list the result in Table~\ref{tab:ablation} (b). The model with ResBlock has better denoising accuracy. Considering the balance between the denoising performance and computation cost, we employ NAFBlock as the basic unit of image encoders.

\textbf{The components of transformer block.} To illustrate the effectiveness of the NRCA module and SVFF module, we first use the vanilla transformer block as the baseline, and then replace the corresponding components with our proposed modules. The vanilla transformer adopts $1\times1$ convolution layer to generate query/key, and directly computes the pixel-level attention map. As shown in Table~\ref{tab:ablation} (c), both of two modules contribute to the improvements of denoising performance. Fig.~\ref{fig:Compare_ablation} visually validates the importance of the two modules. The NRCA module can accurately match the correspondence between two input images (Please refer to Fig.~\ref{fig:NRCA_feature}), and avoid introducing artifacts into the denoised images. The SVFF module enhances structures and edges of the denoised images by regulating the fusion weights of two features.

\textbf{Different attention map propagation strategies.} We replace the neighborhood self-attention layers with $3\times3$ convolution layers, and list the results in Table~\ref{tab:ablation} (d). Though convolution also has the ability to aggregate local information, neighborhood self-attention can enforce the network to capture self-similarity within the attention map, thus achieving higher denoising accuracy.

\textbf{Different feature fusion strategies.} We explore various feature fusion strategies to demonstrate the effectiveness of our SVFF module. By comparing `Add', `Concat', and `Attention' in Table~\ref{tab:ablation} (e), we find that our SVFF module obtains 0.14dB, 0.09dB, and 0.08dB PSNR gains, while having similar learnable parameters. Experiments clearly verify the excellent performance of our selectively feature fusion strategy.

\textbf{The window size of neighborhood self-attention in the attention map propagation.} The influence of window size $k$ is ablated in Table~\ref{tab:ablation} (f). With the increase of $k$, the receptive field of attention map propagation expands. Considering that the computation cost has a quadratic complexity in terms of $k$, we set $k$ to 5.

\textbf{The number of transformer blocks.} We vary $L$ from 1 to 4, and show the results in Table~\ref{tab:ablation} (g). As show in the results, more transformer blocks contribute to better denoising performance, benefiting from multiple feature interaction between the input two images. To balance the computational cost and the performance, we set $L$ to 3 for SGDFormer{$^\dagger$}.

\subsection{Applications to Guided Depth Super-Resolution}

\begin{table}
	\caption{Quantitative comparison of stereo guided depth super-resolution ($\times 8$) on the FlyingThings3D dataset.}
	\renewcommand{\tabcolsep}{8pt}{}
	\renewcommand\arraystretch{1.2}
	\centering
	{
		\begin{tabular}{c|ccc}
			\hline
			Method & NAFSSR~\cite{NAFSSR_CVPRW22} & SANet~\cite{SANet_CVPR23} & SGDFormer (Ours) \\
			\hline
			PSNR $\uparrow$ & 29.42 & 29.28 & 29.74 \\
			\hline
		\end{tabular}
	}
	\label{tab:comparison_SRX8}
\end{table}

\begin{figure}[t]
	\centering
	\includegraphics[scale=0.7] {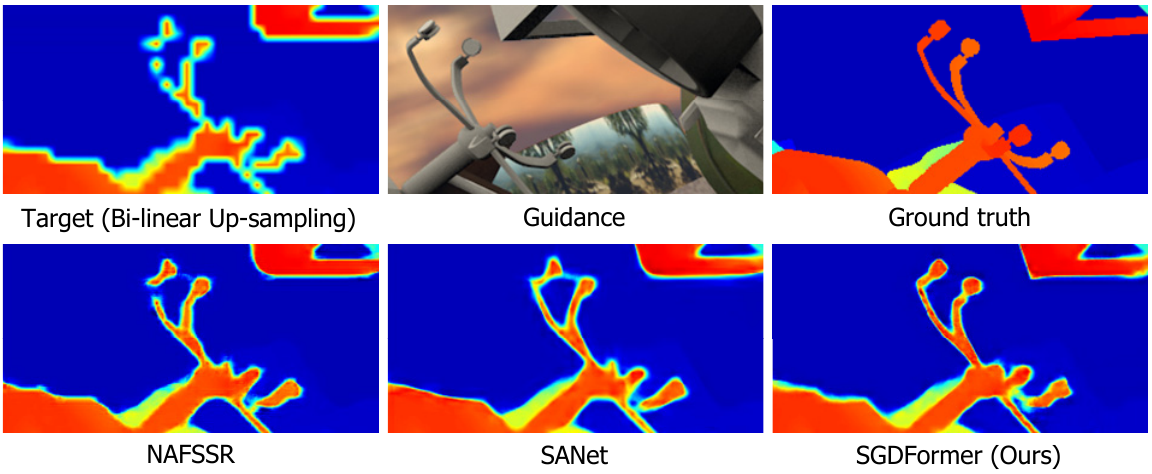}
	\caption{Qualitative comparison of stereo guided depth super-resolution ($\times 8$) on the FlyingThings3D dataset.}
	\label{fig:Res_Fly3d_X8_0}
\end{figure}

Our SGDFormer can also handle other unaligned cross-model guided restoration tasks. We conduct stereo guided depth super-resolution ($\times 8$) on the FlyingThings3D dataset~\cite{Flyingthings3d_CVPR16}. Specifically, we use the right-view RGB image to guide the super-resolution of the left-view depth image. We select 3,986 image pairs for training and 478 image pairs for testing. We compare our SGDFormer with NAFSSR~\cite{NAFSSR_CVPRW22} and SANet~\cite{SANet_CVPR23}. For NAFSSR, the parameters of the NAFBlocks of two image encoders are not shared due to the modality difference, and the stereo cross attention module is used to aggregate RGB features to enhance depth features. The quantitative results in Table~\ref{tab:comparison_SRX8} demonstrate that our SGDFormer achieves the best performance. Fig.~\ref{fig:Res_Fly3d_X8_0} displays an example from the test dataset. Despite significant modality gaps between the two input images, our SGDFormer can still recover fine structures and sharp edges.

\section{Conclusions}
\label{sec:conclusions}

In this paper, we have proposed a one-stage transformer-based architecture, named SGDFormer, for cross-spectral stereo image guided denoising. To better utilize the information of the guidance image, our SGDFormer directly models the correspondence between two images and then performs feature fusion within a unified network, without explicit image registration or aligned guidance image generation. To achieve this, we introduce two neural blocks, \ie the noise-robust cross-attention (NRCA) module and the spatially variant feature fusion (SVFF) module. The NRCA module adopts the attention mechanism in a coarse-to-fine manner, and the SVFF module further enhances salient structures and suppress harmful artifacts. Experiments show that our SGDFormer outperforms previous state-of-the-art approaches. Moreover, our SGDFormer has the potential to cope with other unaligned guided restoration tasks such as guided depth super-resolution.

\section*{Acknowledgment}

This work was supported in part by the National Key Research and Development Program of China under grant 2023YFB3209800, in part by the Natural Science Foundation of Zhejiang Province under grant D24F020006, in part by the National Natural Science Foundation of China under grant 62301484, and in part by the Jinhua Science and Technology Bureau Project.

\bibliographystyle{elsarticle-num} 
\bibliography{ref.bib}

\end{document}